\documentclass{article}
\usepackage{spconf,amsmath,graphicx,hyperref}
\usepackage[numbers]{natbib}
\usepackage{multirow}
\usepackage{subcaption}

\title{Harmonizing Spectral Evolution in Conditional Flow Matching for TTS}
\name{
Isha Pandey$^{\star}$ \qquad
Varad Deshpande$^{\star}$ \qquad
Abhijat Bharadwaj$^{\star}$ \qquad
Ganesh Ramakrishnan
\thanks{$^\star$ These authors contributed equally to this work.}
}

\address{
\emph{Indian Institute of Technology Bombay} \\
\small{\{iishapandey, varadspd, bharadwaj.abhijat, ganramkr\}@iitb.ac.in}
}

\begin{document}
%
\maketitle
\begin{abstract}
Conditional Flow Matching (CFM) models for text-to-speech (TTS) suffer from incoherent frequency evolution during inference. While similar spectral imbalances are addressed in diffusion models for other domains, those generic solutions fail to generalize to the inherently uncoordinated acoustic dynamics of CFM. We demonstrate that this issue can be effectively mitigated by introducing a novel training-free frequency-selective boosting strategy. Using the Discrete Wavelet Transform (DWT), our method dynamically modulates mel-spectrogram sub-bands during ODE integration, synchronizing spectral development by penalizing aggressive low-frequency growth and boosting lagging high-frequency details. Validated across diverse architectures (Matcha-TTS, F5-TTS, IndicF5), our approach reduces the required Number of Function Evaluations (NFE) from 32 to 26 and improves Fréchet Audio Distance (FAD) by up to 61\%, all without compromising mean opinion scores, speaker similarity, and speech intelligibility.
\end{abstract}

\section{Introduction}
\label{sec:intro}

Conditional Flow Matching (CFM)~\citep{cfm_paper} has emerged as a powerful framework for text-to-speech (TTS) synthesis. It learns a vector field $v_\theta$ that transports samples from a simple prior $\mathbf{x}_0 \sim p_0$ to the data distribution by solving the ordinary differential equation (ODE) $\frac{d\mathbf{x}(t)}{dt} = v_{\theta}(x_t, t)$~\citep{cnf}, commonly along efficient Optimal-Transport paths. At inference, the ODE is discretized into a finite number of function evaluations (NFEs). Existing systems use different integration strategies: Voicebox~\citep{voicebox} relies on mid-point ODE solvers, Matcha-TTS~\citep{mehta2024matcha} employs fixed-step Euler integration, whereas F5-TTS~\citep{f5tts} uses Sway Sampling to allocate steps non-uniformly along the trajectory. These approaches primarily optimize temporal discretization; comparatively little attention has been paid to how acoustic frequency components develop during integration.

\begin{figure}[t]
\centering
\includegraphics[width=\linewidth]{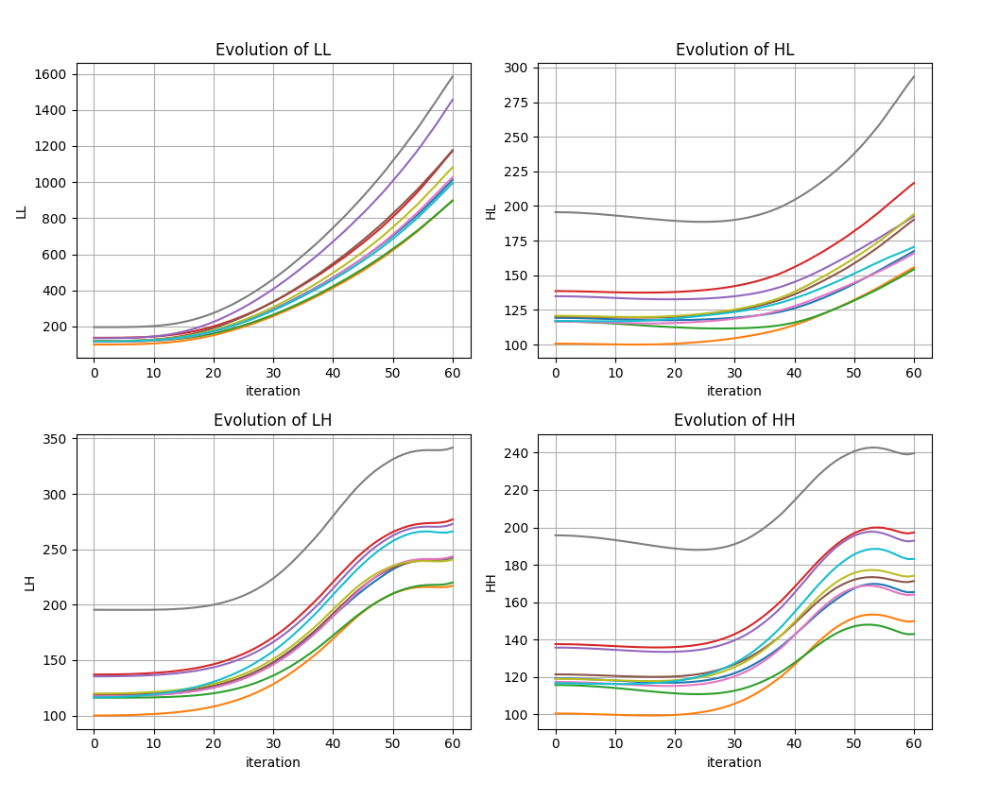}
\caption{Evolution of DWT sub-band energy ($\ell_2$-norm) across F5-TTS inference steps.}
\label{fig:subband-evol-f5tts}
\end{figure}

\begin{figure*}[t]
    \centering
    \captionsetup[subfigure]{font=scriptsize,skip=2pt}

    \begin{subfigure}[t]{0.32\textwidth}
        \centering
        \includegraphics[width=\linewidth]
        {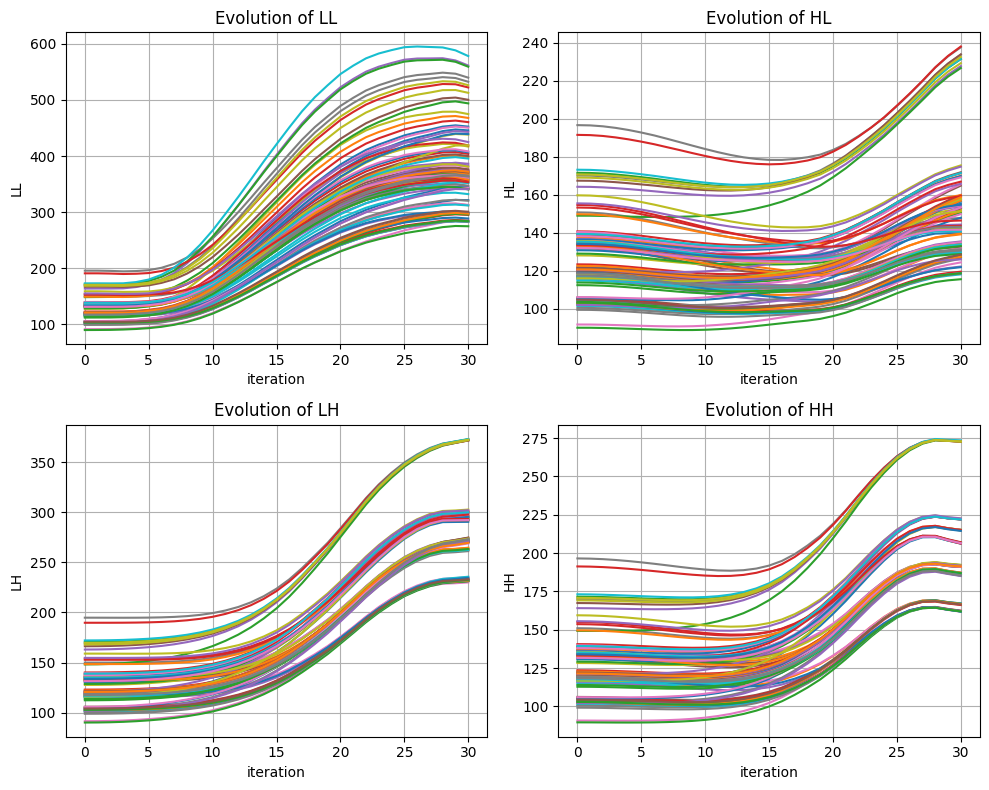}
        \caption{F5-TTS with FSB.}
        \label{fig:f5tts_reweight}
    \end{subfigure}
    \hfill
    \begin{subfigure}[t]{0.32\textwidth}
        \centering
        \includegraphics[width=\linewidth]
        {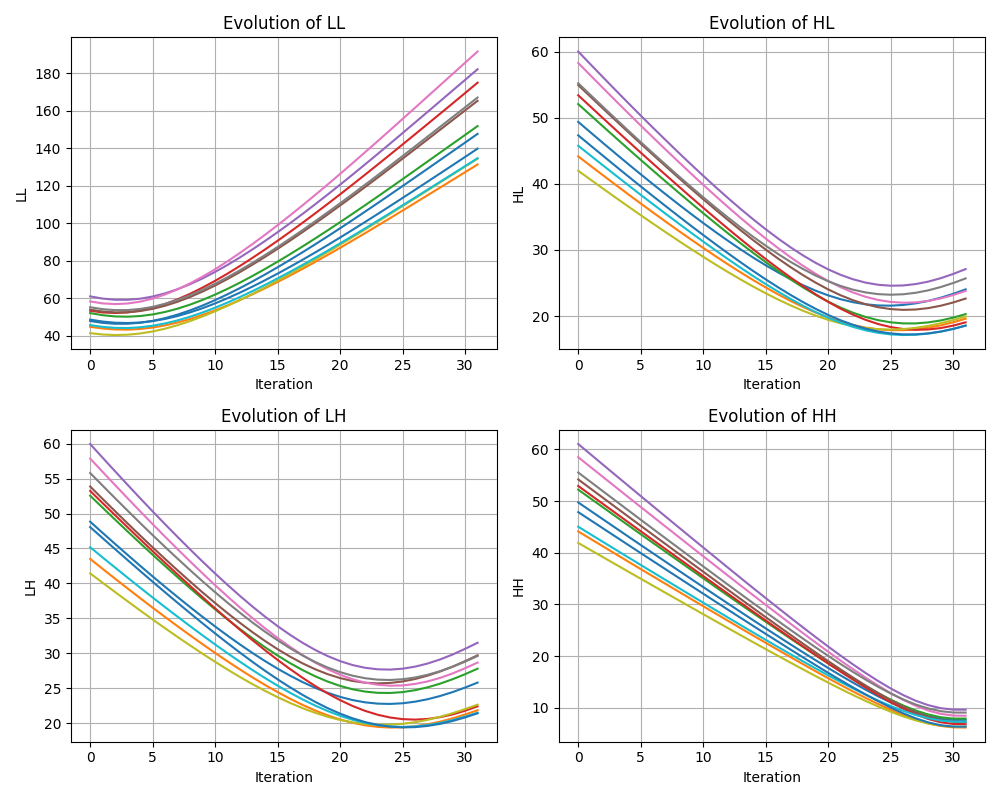}
        \caption{Matcha-TTS baseline.}
        \label{fig:matcha_og}
    \end{subfigure}
    \hfill
    \begin{subfigure}[t]{0.32\textwidth}
        \centering
        \includegraphics[width=\linewidth]
        {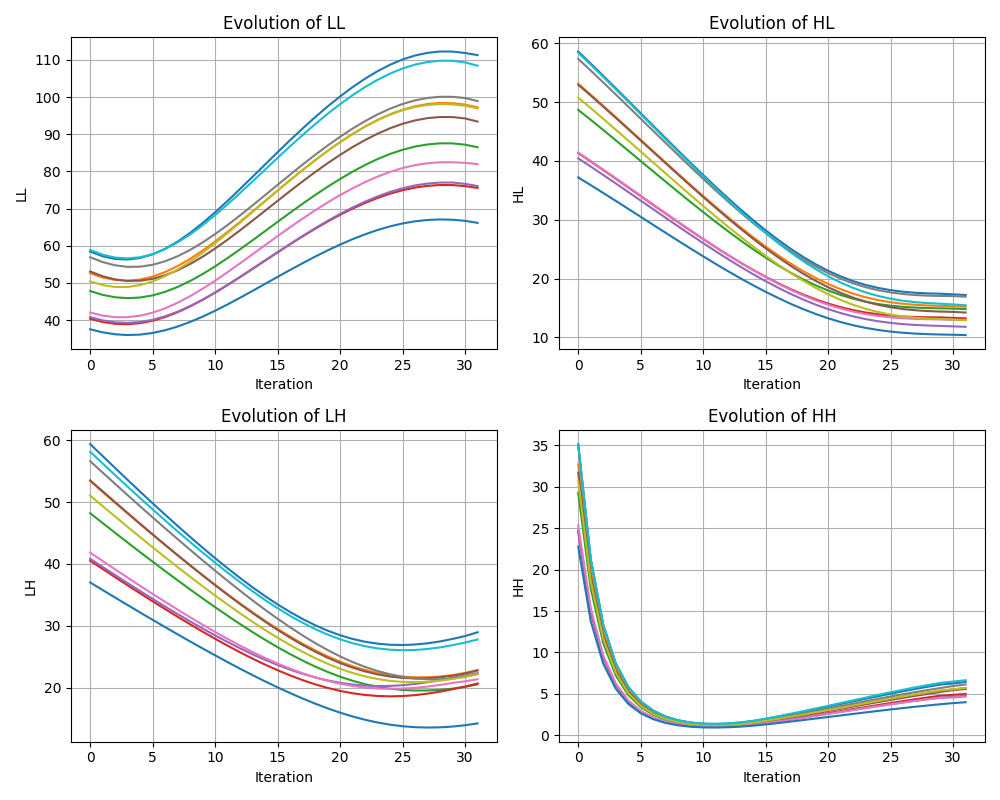}
        \caption{Matcha-TTS with FSB.}
        \label{fig:matcha_rew}
    \end{subfigure}

    \caption{DWT sub-band norm evolution: (a) reweighted F5-TTS, (b) baseline Matcha-TTS,
    and (c) reweighted Matcha-TTS.}
    \label{fig:subband_trajectories}
\end{figure*}

Despite recent innovations, we identify a shared limitation of CFM-based acoustic models: the incoherent evolution of frequency components during ODE integration. As established by \citep{spectral-diffusion}, continuous-time generative models are expected to exhibit a \textit{frequency evolution} property, recovering broad low-frequency structures before transitioning to high-frequency details. However, neural networks possess an intrinsic \textit{frequency bias}~\citep{rahaman2019spectral, spectral-diffusion}: parameter-constrained networks fail to follow this spectral trajectory, leaving them unable to adequately recover high-frequency components towards the end of generation. Following prior spectral analyses~\citep{spectral-diffusion,masf}, we apply the Discrete Wavelet Transform (DWT)~\citep{intro-to-wavelets} to intermediate mel-spectrograms $\mathbf{x}_t$. Across both fixed-step Matcha-TTS and Sway-sampled F5-TTS, the approximation sub-band continues to grow while detail sub-bands develop more slowly (Figures~\ref{fig:subband-evol-f5tts} and~\ref{fig:matcha_og}), revealing a persistent imbalance not resolved by the choice of ODE time grid alone.

To address this imbalance, we introduce \textbf{Frequency-Selective Boosting (FSB)}, a lightweight, training-free inference strategy that decomposes each intermediate mel-spectrogram into DWT sub-bands and applies step-dependent reweighting. FSB suppresses excessive approximation-band growth while boosting lagging detail components, effectively acting as a corrective for the known frequency bias of neural networks rather than an arbitrary spectral intervention. Unlike constant spectral corrections, its modulation strength adapts to the integration stage; moreover, linear schedules from image-generation methods such as MASF~\citep{masf} do not consistently transfer to CFM-based TTS (Table~\ref{tab:schedule_ablation}). The schedules require a one-time calibration analogous to standard inference hyperparameters (NFE count, guidance scale) and can transfer across related architectures and languages. Audio samples are available on our anonymized demo page.\footnote{\url{https://anonymous.4open.science/w/boosed-cfm-demo/}}

This work lays the analytical and empirical foundations for spectral-aware ODE integration in CFM-TTS, establishing that principled sub-band coordination yields consistent gains across architectures and languages. Our contributions are: (1) we identify and quantify incoherent DWT sub-band evolution in CFM-TTS, showing that generic constant-strength or linearly scheduled interventions fail to correct it; and (2) we propose FSB, a training-free reweighting strategy that improves spectral convergence, quality, and efficiency across architectures and languages.

\section{Methodology}
\label{sec:methodology}

\subsection{Sub-band Evolution in CFM-based TTS}
\label{subsec:problem_statement}

At each ODE evaluation, we apply a two-dimensional DWT to the intermediate mel-spectrogram $\mathbf{x}_t$. This produces an approximation sub-band, LL, and three detail sub-bands, LH, HL, and HH. We analyze their trajectories using the $\ell^2$-norm of the corresponding DWT coefficients.

Figures~\ref{fig:subband-evol-f5tts} and~\ref{fig:matcha_og} show that these sub-bands do not develop synchronously. In F5-TTS, LL grows rapidly throughout inference. Although LH and HH begin to stabilize near the end of the trajectory, LL and HL continue to increase. Matcha-TTS exhibits a different profile, characterized by near-linear LL growth and delayed development of the detail sub-bands. Thus, the imbalance persists across both adaptive Sway Sampling and fixed-step Euler integration.

We refer to this behavior as \emph{incoherent sub-band evolution}: the approximation and detail components develop at different rates, with dominant components continuing to grow after other bands have largely stabilized. Our experiments in Section~\ref{sec:results_analysis} show that coordinating sub-band development improves synthesis quality and allows perceptual convergence with fewer function evaluations.

\subsection{Frequency-Selective Boosting}
\label{subsec:our_framework}

\begin{figure}[t]
    \centering
    \includegraphics[
        width=0.82\linewidth,
        trim=0 0 0 0,
        clip
    ]{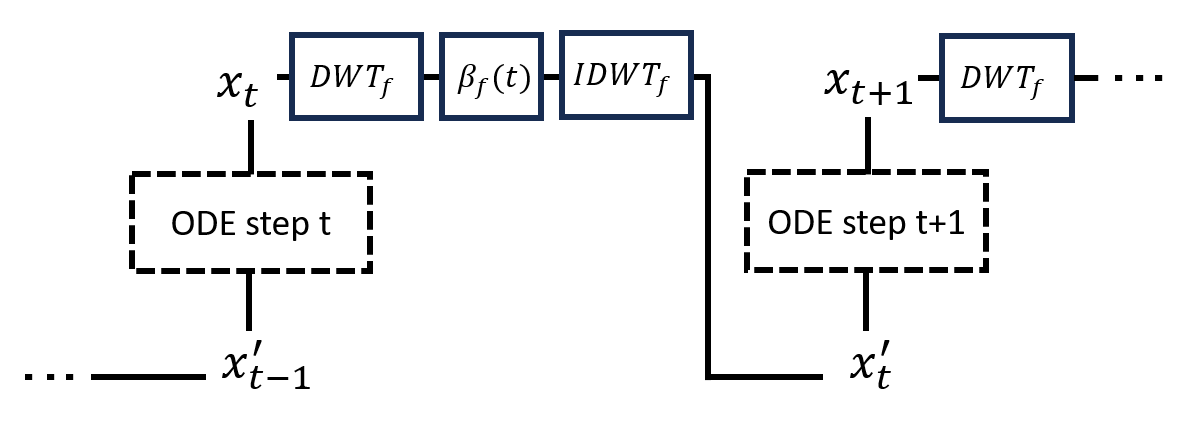}
    \caption{Frequency-selective reweighting process.}
    \label{fig:reweighting-process}
\end{figure}

To address incoherent sub-band evolution, FSB dynamically redistributes the relative contribution of DWT sub-bands as ODE integration progresses. Let $c_f(t)$ denote the observed norm trajectory of sub-band $f$. We define its reweighting coefficient as a clipped, scaled inverse of this trajectory:
\begin{equation}
\label{eq:betagen}
\begin{split}
\beta_f(t;c)
=
\min\left(
\delta_{clip},
\frac{1}{\delta_p+\delta_q c_f(t)}
\right).
\end{split}
\end{equation}
At each ODE evaluation, the intermediate mel-spectrogram $x_t$ is decomposed into its DWT sub-bands, each selected component is scaled by $\beta_f(t)$, and the modified state is reconstructed using the inverse DWT (see Figure~\ref{fig:reweighting-process}):

\begin{equation}
\label{eq:fsb_reweight}
    x'_t
    =
    \mathrm{IDWT}\left(
    \left(
    \beta_f(t)\,\mathrm{DWT}_f(x_t)
    \right)_{f\in\mathcal{F}}
    \right).
\end{equation}
Here, $\mathcal{F}=\{\mathrm{LL},\mathrm{HL},\mathrm{LH},\mathrm{HH}\}$, $\delta_p$ stabilizes the denominator, $\delta_q$ controls the sensitivity of the correction to the observed sub-band energy, and $\delta_{clip}$ prevents excessively large reweighting coefficients. Equation~\ref{eq:betagen} therefore suppresses aggressively growing components while increasing the relative contribution of energy-deficient components. The smooth schedules and clipping operation also prevent abrupt perturbations of the learned ODE trajectory.

For each model, the norm trajectory $c_f(t)$ is represented using a compact family of exponential, linear, or polynomial functions. The corresponding model-specific schedules presented below are low-dimensional approximations to the general inverse reweighting in Equation~\ref{eq:betagen}.

\noindent\textbf{Schedule calibration.}
The schedules are calibrated once from aggregate statistics: we estimate average sub-band growth curves from a small calibration set, select a matching function family, and grid-search $2$--$4$ scalar parameters per modulated sub-band on validation FAD and WER. The schedule is then fixed for all inference---a burden comparable to tuning NFE count or guidance scale. Crucially, IndicF5 reuses the F5-TTS schedule unchanged for Tamil and Bengali (Section~\ref{subsec:crosslingual}).

\noindent\textbf{F5-TTS reweighting.}
The LL sub-band in F5-TTS grows super-linearly (Figure~\ref{fig:subband-evol-f5tts}). We attenuate it using an exponential schedule:
\begin{equation}
\label{eq:llf5}
\begin{split}
\beta_{LL}^{N}(t)
=
\exp\left(-\alpha(N)t\right),
\qquad
\alpha(N)=\frac{\alpha_0}{q(N)}.
\end{split}
\end{equation}
Here, $N$ denotes the total NFE and $\alpha(N)$ controls the attenuation rate. Because the step index $t\in{1,\ldots,N}$ spans different ranges for different $N$, $q(N)$ compensates for this variation. We calibrate $\alpha(N)$ at $\mathcal{S}={10,20,30,45,60}$ and fit a quadratic $q(N)$ across these points, producing an NFE-dependent parameterization that generalizes without a separate search for every $N$. Since HL also continues growing during the final evaluations, it is modulated using the same functional form. LH and HH remain stable and are left unchanged.

\noindent\textbf{Matcha-TTS reweighting.}
Matcha-TTS exhibits a milder, approximately linear imbalance (Figure~\ref{fig:matcha_og}). Small linear corrections to LH and HL suffice:
\begin{equation}
\label{eq:matcha_linear}
\begin{split}
\beta_{LH}(t)=1-b_{LH}\frac{t}{N},
\qquad
\beta_{HL}(t)=1-b_{HL}\frac{t}{N}.
\end{split}
\end{equation}
Here, $b_{LH}$ and $b_{HL}$ control the correction strengths. LL and HH are left unchanged. The resulting trajectories are shown in Figures~\ref{fig:f5tts_reweight} and~\ref{fig:matcha_rew}.

Overall, FSB retains the same DWT-based mechanism across architectures while adapting only its low-dimensional schedule to the observed trajectory profile. Once calibrated, the schedule remains fixed across utterances, requires no model retraining, and introduces only DWT and inverse-DWT operations during inference.

\section{Experimental Setup}
\label{sec:exp_setup}
\noindent\textbf{Model Specifications.}
We use the F5-TTS checkpoint\footnote{\url{https://github.com/SWivid/F5-TTS}} for English and IndicF5~\citep{AI4Bharat_IndicF5_2025} for Indic languages, both $\approx$350M parameters. For Matcha-TTS, we use its checkpoint\footnote{\url{https://github.com/shivammehta25/Matcha-TTS}} trained on the single-speaker LJSpeech dataset~\citep{ljspeech17}.

\noindent\textbf{Evaluation Datasets.}
F5-TTS is evaluated on LibriTTS-clean and -other ($\approx$5k samples each)~\citep{zen2019libritts}. Bengali and Tamil are drawn from IndicVoices-r~\citep{ai4bharat2024indicvoicesr} ($\approx$300 samples). Matcha-TTS is evaluated on LibriSpeech test text ($\approx$400 samples).

\noindent\textbf{Evaluation Metrics.}
Intelligibility: WER and CER using Whisper Large v3~\citep{radford2022whisper} for English and Indic Whisper~\citep{bhogale2023vistaar} for Tamil and Bengali. Quality: Fr\'echet Audio Distance (FAD)~\citep{fadpaper} over VGGish embeddings~\citep{vggish}, UTMOS~\citep{saeki2022utmos} for English, IndicMOS~\citep{udupa2024indicmos} for Indic languages, and subjective MOS. Speaker Similarity Score (SSS): ECAPA-TDNN~\citep{desplanques2020ecapa}. Efficiency: average runtime and Real-Time Factor (RTF).

\section{Results and Analysis}
\label{sec:results_analysis}

\subsection{Quality--Efficiency Trade-off}
With FSB, DWT sub-band norms begin to plateau around the 26th evaluation of a 32-step trajectory (Figures~\ref{fig:f5tts_reweight},~\ref{fig:matcha_rew}), suggesting diminishing spectral changes during the remaining steps. Table~\ref{tab:poc_trajectory} evaluates whether 26 NFEs with FSB can match 32-NFE quality on 1,000 LibriTTS-clean utterances on F5-TTS. We also report cutoff-26/32 (the intermediate state at step 26 of a 32-step run) as a convergence diagnostic.

Full 26-NFE FSB reduces FAD from 1.04 to 0.51 relative to the 26-NFE baseline and outperforms the 32-NFE baseline on all metrics. Extending FSB to 32 NFE yields only marginal further gains, supporting 26 NFE as the preferred operating point. We include Mel-EQ as a non-wavelet baseline: its modest gains confirm that static spectral normalization does not replace time-dependent sub-band coordination. On an NVIDIA H200, 26-NFE FSB reduces average runtime by 16.4\% (1.119\,s vs.\ 1.339\,s) and RTF by 15.9\%.

\begin{table}[t]
\centering
\small
\setlength{\tabcolsep}{4.2pt}
\begin{tabular}{llccc}
\hline
\textbf{Trajectory} & \textbf{Method}
& \textbf{WER $\downarrow$}
& \textbf{CER $\downarrow$}
& \textbf{FAD $\downarrow$} \\
\hline

\multirow{3}{*}{Full 32 NFE}
  & Baseline   & 17.45 & 6.32 & 1.10 \\
  & Mel-EQ     & 17.31 & 6.28 & 1.08 \\
  & FSB (ours) & \textbf{15.05} & \textbf{5.91} & \textbf{0.93} \\
\hline

\multirow{3}{*}{Full 26 NFE}
  & Baseline   & 18.40 & 7.02 & 1.04 \\
  & Mel-EQ     & 17.34 & 6.33 & 0.98 \\
  & FSB (ours) & \textbf{15.50} & \textbf{6.01} & \textbf{0.51} \\
\hline

\multirow{3}{*}{Cutoff-26/32}
  & Baseline   & 18.08 & 7.24 & 7.59 \\
  & Mel-EQ     & 16.59 & 7.01 & 7.32 \\
  & FSB (ours) & \textbf{16.00} & \textbf{6.14} & \textbf{4.66} \\

\hline
\end{tabular}

\caption{Frequency-domain interventions on F5-TTS.}
\label{tab:poc_trajectory}
\end{table}

\subsection{Cross-model and Perceptual Evaluation}
Table~\ref{tab:cross_model} evaluates complete trajectories on full LibriTTS splits for F5-TTS and on LibriSpeech for Matcha-TTS. Relative to matched 26-NFE baselines, FSB reduces FAD by 48\% on LibriTTS-clean, 52\% on LibriTTS-other, and 61\% on Matcha-TTS, while WER/CER and speaker similarity remain close to the 32-NFE references. Predicted MOS recovers most of the degradation from ordinary 26-NFE inference, exceeding the 32-NFE Matcha-TTS score (4.37 vs.\ 4.31).

A human MOS pilot on Matcha-TTS (Table~\ref{tab:human_mos}; 16 raters, 24 samples each using total of 100 samples) corroborates the objective metrics: 26-step FSB leads on Intelligibility and Perceptual Quality and is tied with the 32-step baseline on Naturalness (3.80 vs.\ 3.81).

\begin{table}[t]
\centering
\scriptsize
\setlength{\tabcolsep}{2.7pt}
\resizebox{\columnwidth}{!}{%
\begin{tabular}{lcccccc}
\hline
\textbf{Method}
& \textbf{NFE}
& \textbf{FAD $\downarrow$}
& \textbf{WER $\downarrow$}
& \textbf{CER $\downarrow$}
& \textbf{UTMOS $\uparrow$}
& \textbf{SSS $\uparrow$} \\
\hline

\multicolumn{7}{l}{\textit{F5-TTS on LibriTTS-clean}} \\
Baseline
& 32 & 0.99 & \textbf{6.63} & \textbf{3.76}
& \textbf{3.93} & \underline{0.76} \\
Baseline
& 26 & \underline{0.98} & 6.71 & \underline{3.80}
& 3.05 & 0.74 \\
FSB (ours)
& 26 & \textbf{0.51} & \underline{6.70} & \underline{3.80}
& \underline{3.76} & \textbf{0.77} \\
\hline

\multicolumn{7}{l}{\textit{F5-TTS on LibriTTS-other}} \\
Baseline
& 32 & \underline{1.48} & \underline{6.61} & \textbf{3.28}
& \textbf{3.64} & \underline{0.72} \\
Baseline
& 26 & 1.49 & 6.78 & 3.45
& 2.69 & 0.71 \\
FSB (ours)
& 26 & \textbf{0.72} & \textbf{6.60} & \underline{3.30}
& \underline{3.54} & \textbf{0.73} \\
\hline

\multicolumn{7}{l}{\textit{Matcha-TTS on LibriSpeech}} \\
Baseline
& 32 & \underline{0.49} & \textbf{1.90} & \textbf{0.70}
& \underline{4.31} & -- \\
Baseline
& 26 & 1.15 & 2.30 & 0.92
& 3.63 & -- \\
FSB (ours)
& 26 & \textbf{0.45} & \underline{2.13} & \underline{0.84}
& \textbf{4.37} & -- \\
\hline
\end{tabular}%
}
\caption{Cross-model evaluation on full ODE trajectories.}
\label{tab:cross_model}
\end{table}

\begin{table}[t]
\centering
\scriptsize
\setlength{\tabcolsep}{3.1pt}
\resizebox{\columnwidth}{!}{%
\begin{tabular}{lccc}
\hline
\textbf{Setting} & \textbf{Naturalness}
& \textbf{Intelligibility} & \textbf{Perceptual quality} \\
\hline
Baseline, full 32      & \textbf{3.81} & \underline{4.43} & \underline{4.34} \\
Baseline, cutoff-26/32 & 3.51 & 4.09 & 3.69 \\
FSB, full 26 (ours)    & \underline{3.80} & \textbf{4.45} & \textbf{4.50} \\
\hline
\end{tabular}%
}
\caption{Matcha-TTS human evaluation.}
\label{tab:human_mos}
\end{table}


\begin{table}[t]
\centering
\scriptsize
\setlength{\tabcolsep}{2.7pt}
\resizebox{\columnwidth}{!}{%
\begin{tabular}{llccc}
\hline
\textbf{Dataset}
& \textbf{LL schedule}
& \textbf{FAD $\downarrow$}
& \textbf{WER $\downarrow$}
& \textbf{CER $\downarrow$} \\
\hline
\multirow{4}{*}{LibriTTS-clean}
& Baseline    & 12.93 & \textbf{6.7} & \textbf{3.7} \\
& Exponential & \textbf{10.78} & 7.2 & 4.7 \\
& Linear      & 13.15 & 8.1 & 5.3 \\
& Polynomial  & 17.29 & 19.8 & 13.3 \\
\hline
\multirow{4}{*}{LibriTTS-other}
& Baseline    & 9.03 & 8.3 & 5.1 \\
& Exponential & \textbf{8.82} & \textbf{8.2} & \textbf{4.9} \\
& Linear      & 10.78 & 8.3 & 5.0 \\
& Polynomial  & 14.36 & 43.1 & 30.4 \\
\hline
\end{tabular}%
}
\caption{
Ablation of LL-subband reweighting schedule (cutoff-26/32). Other subband coefficients held fixed.
}
\label{tab:schedule_ablation}
\end{table}

\subsection{Ablation: Reweighting Schedules}
\label{subsec:schedule_ablation}
Table~\ref{tab:schedule_ablation} compares exponential, linear, and degree-3 polynomial schedules for the LL sub-band under the cutoff-26/32 diagnostic. The exponential schedule achieves the lowest FAD on both splits; the linear schedule is competitive, indicating robustness to minor schedule variations; the polynomial schedule fails catastrophically, confirming that the exponential form best counteracts super-linear low-frequency growth.

\subsection{Ablation: Cross-lingual Transfer}
\label{subsec:crosslingual}

Using the cutoff-26/32 diagnostic on IndicF5, FSB halves FAD for both Tamil (7.58$\to$4.37) and Bengali (9.30$\to$4.32) while improving predicted MOS (2.24$\to$2.30 and 2.07$\to$2.26) and speaker similarity (0.78$\to$0.81 and 0.75$\to$0.78). ASR metrics are mixed, though the ground-truth recordings themselves produce WER of 62--65\%, making ASR unreliable here; the distributional and perceptual metrics consistently favor FSB. Crucially, IndicF5 reuses the F5-TTS schedule without modification, demonstrating cross-lingual transfer within the same architecture family.

\subsection{Trajectory Stability}

Modifying the state between ODE evaluations risks abrupt corrections or
compounding drift. We diagnose this using finite-difference acceleration
$\hat{a}_i$ and the log-linear divergence rate $r$ from fitting
$\log\|x_t^{\mathrm{mod}}-x_t^{\mathrm{base}}\|_2=rt+c$.
As shown in Table~\ref{tab:trajectory_stability}, FSB (LL+HL) maintains
baseline-level mean acceleration ($1.01\times$) while \emph{reducing} peak
acceleration to $0.80\times$, confirming that reweighting does not introduce
corrective spikes. LL-only modulation has slightly lower acceleration but
higher divergence ($r{=}0.089$ vs.\ $0.049$), indicating that co-modulating
HL stabilizes the trajectory by keeping detail sub-bands synchronized with
the attenuated approximation band. In contrast, Pre-Emphasis increases
acceleration by $>$6$\times$ with strongly exponential divergence
($R^2{=}0.971$), consistent with its severe intelligibility degradation.


\begin{table}[t]
\centering
\scriptsize
\setlength{\tabcolsep}{3.4pt}
\resizebox{\columnwidth}{!}{%
\begin{tabular}{lccc}
\hline
\textbf{Method}
& \textbf{Mean $\|\hat{a}\|$}
& \textbf{Max $\|\hat{a}\|$}
& $\boldsymbol{r\;(R^2)}$ \\
\hline
Baseline
& $1.00{\times}$
& $1.00{\times}$
& -- \\

FSB: LL+HL
& $1.01{\times}$
& $0.80{\times}$
& $0.049\;(0.787)$ \\

DWT: LL only
& $0.93{\times}$
& $0.77{\times}$
& $0.089\;(0.698)$ \\

Mel-EQ
& $1.03{\times}$
& $1.03{\times}$
& $0.042\;(0.786)$ \\

Pre-Emphasis
& $6.62{\times}$
& $6.35{\times}$
& $0.087\;(0.971)$ \\
\hline
\end{tabular}%
}
\caption{
Trajectory-stability diagnostics (100 LibriTTS-clean samples).
Acceleration normalized by baseline; $r$ is the log-linear divergence rate.
}
\label{tab:trajectory_stability}
\end{table}

\section{Conclusion}
\label{sec:conclusion}
We identified incoherent DWT sub-band evolution as a shared limitation of CFM-based TTS models and proposed Frequency-Selective Boosting (FSB), a training-free, step-dependent reweighting strategy that coordinates spectral development during ODE integration. FSB reduces the required NFE from 32 to 26 and improves FAD by up to 61\% across F5-TTS, Matcha-TTS, and IndicF5, without degrading intelligibility and perceptual quality. Its low-dimensional schedules require only a one-time calibration and transfer across languages. Future work includes automatic schedule estimation and extension to waveform-domain flow models.

\vfill\pagebreak

\bibliographystyle{IEEEbib}
\bibliography{strings}

@ARTICLE{intro-to-wavelets,
  author={Graps, A.},
  journal={IEEE Computational Science and Engineering}, 
  title={An introduction to wavelets}, 
  year={1995},
  volume={2},
  number={2},
  pages={50-61},
  doi={10.1109/99.388960}}

@misc{f5tts,
      title={F5-TTS: A Fairytaler that Fakes Fluent and Faithful Speech with Flow Matching}, 
      author={Yushen Chen and Zhikang Niu and Ziyang Ma and Keqi Deng and Chunhui Wang and Jian Zhao and Kai Yu and Xie Chen},
      year={2024},
      eprint={2410.06885},
      archivePrefix={arXiv},
      primaryClass={eess.AS},
      url={https://arxiv.org/abs/2410.06885}, 
}

@misc{fadpaper,
      title={Fr\'echet Audio Distance: A Metric for Evaluating Music Enhancement Algorithms}, 
      author={Kevin Kilgour and Mauricio Zuluaga and Dominik Roblek and Matthew Sharifi},
      year={2019},
      eprint={1812.08466},
      archivePrefix={arXiv},
      primaryClass={eess.AS},
      url={https://arxiv.org/abs/1812.08466}, 
}

@inproceedings{
voicebox,
title={Voicebox: Text-Guided Multilingual Universal Speech Generation at Scale},
author={Matthew Le and Apoorv Vyas and Bowen Shi and Brian Karrer and Leda Sari and Rashel Moritz and Mary Williamson and Vimal Manohar and Yossi Adi and Jay Mahadeokar and Wei-Ning Hsu},
booktitle={Thirty-seventh Conference on Neural Information Processing Systems},
year={2023},
url={https://openreview.net/forum?id=gzCS252hCO}
}

@inproceedings{
cfm_paper,
title={Flow Matching for Generative Modeling},
author={Yaron Lipman and Ricky T. Q. Chen and Heli Ben-Hamu and Maximilian Nickel and Matthew Le},
booktitle={The Eleventh International Conference on Learning Representations },
year={2023},
url={https://openreview.net/forum?id=PqvMRDCJT9t}
}

@InProceedings{masf,
    author    = {Qian, Yurui and Cai, Qi and Pan, Yingwei and Li, Yehao and Yao, Ting and Sun, Qibin and Mei, Tao},
    title     = {Boosting Diffusion Models with Moving Average Sampling in Frequency Domain},
    booktitle = {Proceedings of the IEEE/CVF Conference on Computer Vision and Pattern Recognition (CVPR)},
    month     = {June},
    year      = {2024},
    pages     = {8911-8920}
}

@inproceedings{cnf,
author = {Chen, Ricky T. Q. and Rubanova, Yulia and Bettencourt, Jesse and Duvenaud, David},
title = {Neural ordinary differential equations},
year = {2018},
publisher = {Curran Associates Inc.},
address = {Red Hook, NY, USA},
booktitle = {Proceedings of the 32nd International Conference on Neural Information Processing Systems},
pages = {6572–6583},
numpages = {12},
location = {Montr\'{e}al, Canada},
series = {NIPS'18}
}

@misc{radford2022whisper,
  doi = {10.48550/ARXIV.2212.04356},
  url = {https://arxiv.org/abs/2212.04356},
  author = {Radford, Alec and Kim, Jong Wook and Xu, Tao and Brockman, Greg and McLeavey, Christine and Sutskever, Ilya},
  title = {Robust Speech Recognition via Large-Scale Weak Supervision},
  publisher = {arXiv},
  year = {2022},
  copyright = {arXiv.org perpetual, non-exclusive license}
}

@INPROCEEDINGS{vggish,
  author={Hershey, Shawn and Chaudhuri, Sourish and Ellis, Daniel P. W. and Gemmeke, Jort F. and Jansen, Aren and Moore, R. Channing and Plakal, Manoj and Platt, Devin and Saurous, Rif A. and Seybold, Bryan and Slaney, Malcolm and Weiss, Ron J. and Wilson, Kevin},
  booktitle={2017 IEEE International Conference on Acoustics, Speech and Signal Processing (ICASSP)}, 
  title={CNN architectures for large-scale audio classification}, 
  year={2017},
  volume={},
  number={},
  pages={131-135},
  doi={10.1109/ICASSP.2017.7952132}}

@INPROCEEDINGS{spectral-diffusion,
  author={Yang, Xingyi and Zhou, Daquan and Feng, Jiashi and Wang, Xinchao},
  booktitle={2023 IEEE/CVF Conference on Computer Vision and Pattern Recognition (CVPR)}, 
  title={Diffusion Probabilistic Model Made Slim}, 
  year={2023},
  volume={},
  number={},
  pages={22552-22562},
  doi={10.1109/CVPR52729.2023.02160}}

@misc{AI4Bharat_IndicF5_2025,
  author       = {Praveen S V and Srija Anand and Soma Siddhartha and Mitesh M. Khapra},
  title        = {IndicF5: High-Quality Text-to-Speech for Indian Languages},
  year         = {2025},
  url          = {https://github.com/AI4Bharat/IndicF5},
}

@article{ai4bharat2024indicvoicesr,
  title={IndicVoices-R: Unlocking a Massive Multilingual Multi-speaker Speech Corpus for Scaling Indian TTS},
  author={Sankar, Ashwin and Anand, Srija and Varadhan, Praveen Srinivasa and Thomas, Sherry and Singal, Mehak and Kumar, Shridhar and Mehendale, Deovrat and Krishana, Aditi and Raju, Giri and Khapra, Mitesh},
  journal={NeurIPS 2024 Datasets and Benchmarks},
  year={2024}
}

@inproceedings{zen2019libritts,
  title        = {LibriTTS: A Corpus Derived from LibriSpeech for Text‐to‐Speech},
  author       = {Heiga Zen and Viet Dang and Rob Clark and Yu Zhang and Ron J. Weiss and Ye Jia and Zhifeng Chen and Yonghui Wu},
  booktitle    = {Proc.\ Interspeech},
  year         = {2019},
  doi          = {10.21437/Interspeech.2019-2441},
}

@inproceedings{mehta2024matcha,
  title={Matcha-TTS: A fast TTS architecture with conditional flow matching},
  author={Mehta, Shivam and Tu, Ruibo and Beskow, Jonas and Sz{\'e}kely, {\'E}va and Henter, Gustav Eje},
  booktitle={ICASSP 2024-2024 IEEE International Conference on Acoustics, Speech and Signal Processing (ICASSP)},
  pages={11341--11345},
  year={2024},
  organization={IEEE}
}

@misc{ljspeech17,
  author       = {Keith Ito and Linda Johnson},
  title        = {The LJ Speech Dataset},
  howpublished = {\url{https://keithito.com/LJ-Speech-Dataset/}},
  year         = 2017
}

@article{bhogale2023vistaar,
  title={Vistaar: Diverse benchmarks and training sets for indian language asr},
  author={Bhogale, Kaushal Santosh and Sundaresan, Sai and Raman, Abhigyan and Javed, Tahir and Khapra, Mitesh M and Kumar, Pratyush},
  journal={arXiv preprint arXiv:2305.15386},
  year={2023}
}

@inproceedings{rahaman2019spectral,
  title={On the spectral bias of neural networks},
  author={Rahaman, Nasim and Baratin, Aristide and Arpit, Devansh and Draxler, Felix and Lin, Min and Hamprecht, Fred and Bengio, Yoshua and Courville, Aaron},
  booktitle={International conference on machine learning},
  pages={5301--5310},
  year={2019},
  organization={PMLR}
}

@inproceedings{saeki2022utmos,
  title={UTMOS: UTokyo-SaruLab System for VoiceMOS Challenge 2022},
  author={Saeki, Takaaki and others},
  booktitle={Proc. Interspeech 2022},
  year={2022}
}

@inproceedings{udupa2024indicmos,
  title={IndicMOS: Multilingual MOS Prediction for 7 Indian Languages},
  author={Udupa, Sathvik and others},
  booktitle={Proc. Interspeech 2024},
  year={2024}
}

@inproceedings{desplanques2020ecapa,
  title={ECAPA-TDNN: Emphasized Channel Attention, Propagation and Aggregation in Time Delay Neural Network},
  author={Desplanques, Brecht and Thienpondt, Jenthe and Demuynck, Kris},
  booktitle={Proc. Interspeech 2020},
  year={2020}
}

\end{document}